\documentclass[letterpaper]{article} 
\usepackage[draft]{aaai2026}  
\usepackage{times}  
\usepackage{helvet}  
\usepackage{courier}  
\usepackage[hyphens]{url}  
\usepackage{graphicx} 
\usepackage{natbib}  
\usepackage{caption} 
\usepackage{algorithm}
\usepackage{algorithmic}

\usepackage{newfloat}
\usepackage{listings}
\DeclareCaptionStyle{ruled}{labelfont=normalfont,labelsep=colon,strut=off} 
\floatstyle{ruled}
\newfloat{listing}{tb}{lst}{}
\floatname{listing}{Listing}
\usepackage{booktabs}
\usepackage{multirow}
\usepackage{microtype}
\usepackage{inconsolata}
\usepackage{tabularx}
\usepackage{graphicx}
\usepackage{multirow}
\usepackage{algorithmic}
\usepackage{arydshln}
\usepackage{amssymb}
\usepackage{listings}
\usepackage{rotating}
\usepackage{algorithm}
\usepackage{amsmath}

\usepackage{hyperref}
\hypersetup{
    colorlinks=true,
    urlcolor=blue,
    linkcolor=blue,
    citecolor=blue,
}
\usepackage{url}

\title{Structured Language Generation Model: Loss Calibration and Formatted Decoding for Robust Structure Prediction and Knowledge Retrieval}

\author{Minho Lee\textsuperscript{1} \quad
    Junghyun Min\textsuperscript{2} \quad 
    Yerang Kim\textsuperscript{3} \quad
    Woochul Lee\textsuperscript \quad
    Yeonsoo Lee\textsuperscript{5}
    \\[1ex]
  \textsuperscript{1}KT Gen AI Lab \quad
  \textsuperscript{2}Georgetown University \quad
  \textsuperscript{3}Korea University \quad 
  \textsuperscript{5}NC AI\\
  \textsuperscript{1}\texttt{minolee@kt.com} \quad\quad
  \textsuperscript{3}\texttt{hs01151116@korea.ac.kr}
}

\begin{document}
\maketitle
\begin{abstract}

Modern generative pre-trained language models excel at open-ended text generation, yet continue to underperform on structure-related tasks such as NER, relation extraction, and semantic role labeling, especially when compared to encoder-only models of similar sizes.
While this gap has been attributed to limited structure knowledge, we hypothesize this is also due to the missing connection between the model’s internal representations of linguistic structure and the output space used during supervised fine-tuning.
We propose the Structured Language Generation Model (SLGM), a model- and task-agnostic framework that reformulates structured prediction as a classification problem through three components: (1) reinforced input formatting with structural cues, (2) loss design, and (3) format-aware decoding that constrains generation to task-valid outputs.
Across 5 tasks and 13 datasets, SLGM substantially improves structure prediction without relying on dataset-specific engineering or additional model parameters, strengthening alignment between the model's internal structure representation and output. It outperforms baseline fine-tuning on models of the same size, achieves comparable performance to much larger models when used with $<$1B parameter models, and acts as a zero-weight adapter that reproduces the benefits of dataset-specific fine-tuning in low-resource settings.

\end{abstract}

\section{Introduction}

Recent advances in pre-trained language models (PLMs) allow computational approaches to more language tasks and problems than ever \citep{devlin-etal-2019-bert, t5}.
Generative PLMs (GLMs) perform strongly on natural language generation \citep{gpt2, bart}, and engineering methods like scaling and post-training methods like instruction tuning and reinforcement learning from human feedback further improve generation quality \citep{gpt3,  flant5, bai2023qwentechnicalreport, openai_chatgpt}.

However, such success has not been mirrored in many natural language understanding (NLU) tasks that require knowledge of syntactic and semantic structure where the scale and post-training associated with GLMs offer little improvement over much smaller, unaligned LMs \citep{zhong2023chatgpt, hu2025improvinglanguageunderstandingcapabilities}.

Beyond benchmark scores, GLMs still struggle to represent linguistic structure compared to their strength in next-token prediction. 
GLMs' ability to retrieve previously seen entities depends on the order of the input \citep{berglund2024the, kitouni2024factorization}.
Their ability to predict syntactic structure remains poor even with in-context learning \citep{bai2023constituency}.
\citet{deepstruct, min-etal-2025-punctuation} show that additional structure pre-training improves performance on structure-related downstream tasks like information extraction.
Table \ref{tab:deepstruct_dataset_ablation} illustrates a similar behavior, where GPT-4 \citep{openai_chatgpt} falls short against smaller models in CoNLL04 joint entity and relation extraction \citep{conll2004}, even when equipped with detailed instructions and examples, echoing similar sentiments from \citet{li2023evaluatingchatgptsinformationextraction, han2024empiricalstudyinformationextraction}, who observe relatively subpar LLM performance in structure- and retrieval-related tasks.

\begin{figure}[t]
    \centering
    \includegraphics[width=\linewidth]{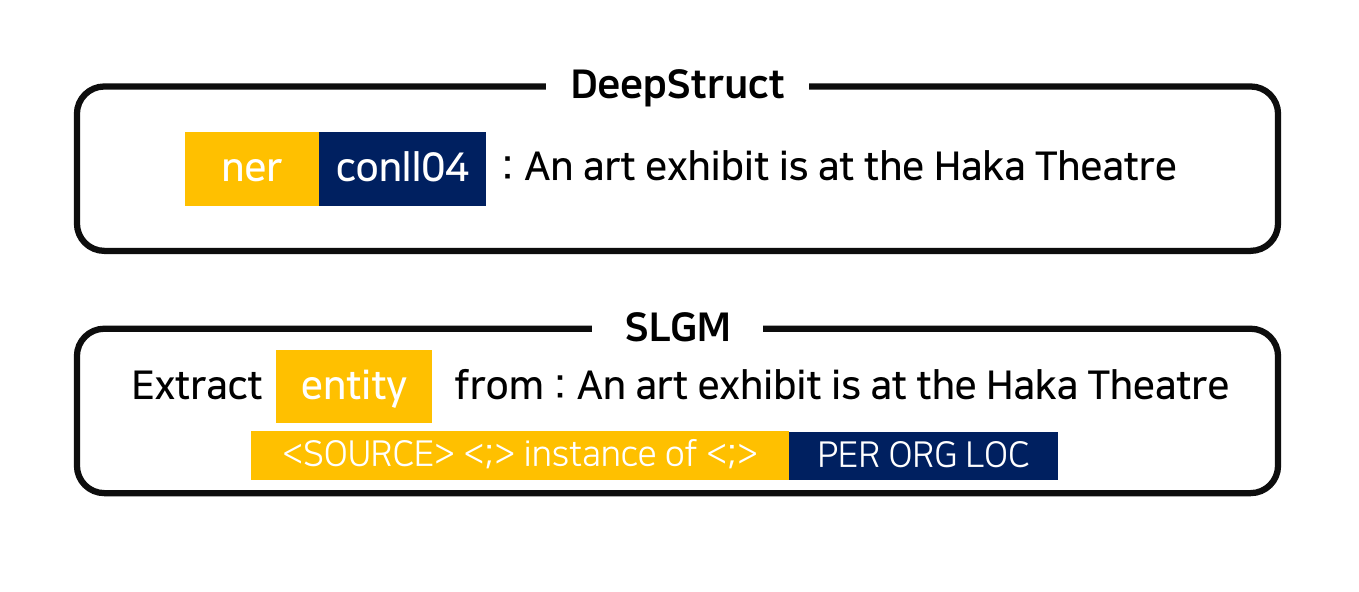}
    \caption{Sample NER example in \textsc{DeepStruct} \citep{deepstruct} and SLGM. The yellow and navy highlight task specific and dataset information, respectively. SLGM breaks down the implicit information into explicit output format and tagset information.}
    \label{fig:intro}
\end{figure}

However, GLMs' lower-than-expected performance on various structure-related tasks that require internal knowledge retrieval, including named entity recognition (NER), relation extraction (RE), semantic role labeling (SRL), intent detection (ID), and dialogue state tracking (DST) may not be entirely attributed to the lack of linguistic structure knowledge in them, as they are capable of syntactically well-formed and semantically coherent text \citep{Herbold2023largescale,Olmedilla2024evaluating,Acciai2025narrative, lauriola-etal-2025-analyzing}. 
We hypothesize that GLMs' comparatively weak performance in structure-related tasks is due to the missing connection between the internal representations of linguistic structure and the output tokens, and test how model- and dataset-agnostic designs like input formatting, loss design, and restrictions in decoding can help retrieve structure information from GLMs more effectively.

In this paper, we present our model- and task-agnostic framework, the Structured Language Generation Model (SLGM), which comprises reinforced input formatting, loss functions complementary to the cross-entropy loss, and format-aware decoding method. An example of its input formatting is illustrated in Figure \ref{fig:intro}.

We show that the SLGM framework is able to bridge the gap between GLMs' ability to capture hierarchical structure in language and their subpar structure prediction performance obtained from vanilla fine-tuning, providing empirical analysis on the effects of additional loss and formatted decoding.
SLGM offers significant performance improvements over baseline models of the same size, and is competitive against other much larger $>$10B models even when used with $<$1B models, achieving the state-of-the-art performance in NER with CoNLL-03 \citep{conll2003} and intent detection with ATIS \citep{atis}.
We also present SLGM as an effective zero-weight adapter that simulates dataset-specific fine-tuning for low-resource environments.

\begin{table}[t!]
\centering
\begin{tabular}{llcc}
\toprule
\textbf{Model} & \textbf{Meta-info} & \textbf{Ent. F1} & \textbf{Rel. F1} \\
\midrule
\multirow{3}{*}{\textsc{DeepStruct}} 
  & +task +dset & 88.4 & 72.8 \\
  & +task +tag  & 81.7 & 40.7 \\
  & +task       & 79.6 & 48.2   \\
\midrule
\multirow{2}{*}{TANL} 
  & +dset   & 90.3 & 70.0 \\
  & -       & 24.0 & 2.8  \\
\midrule
GPT-4* & +task +tag & 57.7 & 34.1 \\
\midrule
\multirow{2}{*}{SLGM} 
  & +task +tag & 71.7 & 27.9 \\
  & +task +tag +ft  & 85.5 & 55.2 \\
\bottomrule
\end{tabular}
\caption{\textsc{DeepStruct} \citep{deepstruct} and TANL \citep{tanl} performance on CoNLL-04 joint entity recognition and relation classification (JER) are better with task information (+task), dataset information (+dset), tagset information (+tag), and degrade without them. GPT-4 \citep{openai_chatgpt} performance on a test subset and SLGM performance with and without finetuning (+ft) as reference; Appendix \ref{appendix:chatgpt-instruction} describes GPT-4 prompt.
}
\label{tab:deepstruct_dataset_ablation}
\end{table}


\section{Structure Prediction and Knowledge Retrieval}

Structure-related tasks that require entity or relation retrieval have originally built on the sequence generation framing of NLP tasks, widely adopted by generative pre-trained models \citep{bart, flant5, glm, t5}.
For greater and more reliable performance, TANL \citep{tanl}, DeepEx \citep{deepex}, and UIE \citep{uie} propose intermediate representations of entity and argument structure within text, which may be comparable to more general representations of meaning \citep[e.g. Abstract Meaning Representations; ][]{banarescu-etal-2013-abstract}, incorporated into downstream tasks with mixed results \citep{wein-opitz-2024-survey, jin-etal-2024-analyzing, min2025doesmeaningbackfireinvestigating}.

Recent work explores how explicit structure modeling can enhance generation or extraction performance.
Large and even commercial language models underperform much smaller fine-tuned models for information extraction and other structure- and retrieval-related tasks even with test-time scaling methods like in-context learning and chain-of-thought reasoning \citep{li2023evaluatingchatgptsinformationextraction, han2024empiricalstudyinformationextraction}.
\citet{uie}, \citet{ deepstruct}, and \citet{min-etal-2025-punctuation} introduce additional structure-related pre-training, with or without supervision, while \citet{wang-etal-2021-automated} concatenate existing pre-trained representations to improve downstream task performance.
In a parallel line of research, \citet{sun-etal-2020-predicate} and \citet{wang-etal-2022-oie} propose the Open Information Expression (OIX) and Open Information Annotation (OIA) frameworks, which represent predicate–argument relations in a unified, lossless format.
These frameworks decouple structural annotation from task-specific modeling, allowing OIE strategies to be reused and adapted efficiently across diverse tasks.
The OIA-based system of \citet{wang-etal-2022-oie} further demonstrates that explicitly modeling predicate–function–argument structures can yield strong adaptability and competitive performance even with limited training data, underscoring the value of explicit structure representations in general-purpose information extraction.


By definition, structured prediction tasks benefit from decoding methods that restrict the set of candidate tokens to choose from.
Such characteristic resembles the copying mechanism \citep{copy_mechanism}, which allows subsequences of source text to be "copied" when generating target text.
This approach has been adopted for summarization \citep{copy_summary, see-etal-2017-get, bottom_up_summarization, force_copy}, dialogue systems \citep{copyt5}, machine translation \citep{maskedgen_mt}, data-to-text generation \citep{force_copy}, and data augmentation for aspect-based sentiment analysis \citep{maskedgen_aspect}.


Another challenge in structured prediction tasks lies in producing outputs with precise formatting. Earlier work often relied on rule-based systems \citep[e.g.][]{rule-based-generation}, which guaranteed stability but lacked the ability to generalize. In contrast, pretrained language models offer broad generalization power, yet controlling their outputs remains difficult \citep{zhang2023control}.
Despite their broad generalization abilities, pretrained language models often fail to reliably adhere to structural or formatting constraints, motivating approaches for constrained decoding and controllable generation \citep{keskar2019ctrlconditionaltransformerlanguage, holtzman2020curious,  lu-etal-2021-neurologic}.
Work in code generation have explored ways to enforce structure during generation, such as constraining decoding with formal grammars or pushdown automata \citep{codepad}, or adjusting the granularity of token generation \citep{codegen_base}. At the same time, large commercial models demonstrate strong capabilities in reliably producing well-structured outputs like code \citep{bai2023qwentechnicalreport,openai_chatgpt, gemmateam2025gemma3technicalreport}, highlighting the potential of combining powerful language modeling with explicit control mechanisms.

Beyond and sentence-level structure prediction, as the role of LLMs as search engines or information retrieval systems grows \citep{lewis2020retrieval, wang2024large}, research in LLMs have also been concerned with structural reliability and effective retrieval \citep{willard2023efficientguidedgenerationlarge, zheng2024sglang, li2025structrag}.
\citet{willard2023efficientguidedgenerationlarge, zheng2024sglang} use a finite state machine \citep[FSM; ][]{McCulloch1943} to guide decoding to generate while following an output format described as a regular expression.
This ensures precise formatting in schemas like JSON, while also restricting the set of candidates to choose from during decoding.

\citet{li2025structrag} propose "structurization" of knowledge scattered throughout a database of documents or corpora before retrieving relevant knowledge and generating a response to a question.
We liken this intermediate knowledge conversion to table-like objects with structure to many intermediate representations of meaning and structure in NLP frameworks \citep{banarescu-etal-2013-abstract, sun-etal-2020-predicate, wang-etal-2022-oie}.
They motivate their framework to human behavior of preferring various forms of structured data to solve reasoning task \citep{sweller1988cognitive, chandler1991cognitive} rather than working with raw text \citep{johnson1983mental, paivio1990mental}.
Such approaches illustrate the importance of explicit control mechanisms for LLM generation and the benefits of "structure-augmented generation" even in today's LLM-powered, retrieval-augmented systems.

\section{SLGM Framework}
\label{sec:Method}


SLGM operates around task and data-specific information. 
For each task and dataset, we establish a predefined tagset and output format, which are then enforced via loss and forced decoding outlined in this section.

\subsection{Task-specific Output Format}
\label{task-specific-format}

Each task or dataset, when framed as a sequence-to-sequence task, has its own output format.
Such format consists of separator tokens and the segments between them, which we call \textit{slots}.
For example, a single output string may contain slot separator tokens (\texttt{<;>}) and an object separator token (\texttt{</>}) to indicate the end of the object, which can be a named entity, a verb with semantic roles, or a head-predicate-tail triple in information extraction.
Each slot can contain a list of strings or one of two special slot tokens - \texttt{<ANY>} and \texttt{<SOURCE>}.
An \texttt{<ANY>} token indicates the model can generate any token, while the \texttt{<SOURCE>} token indicates the model should source from the input.
If a list is given in the slot, the model is to choose from the list of pre-defined strings, which contains tagset information or other placeholder labels.
Before inference, SLGM parses the format string into a format mask tensor, a boolean tensor that has true values on legal token ids for each slot.
This tensor is then used in calculating format losses or disabling illegal tokens in the decoding process, described in the following Sections \ref{sec:format-loss} and \ref{sec:formatted-decoding}.

\subsection{Format Loss}
\label{sec:format-loss}


Traditionally popular cross-entropy loss is calculated over every possible vocabulary defined in the model.
We find this to target an overly broad token space.
Thus, we introduce two complementary losses: a structure loss, and a slot loss.

\paragraph{Structure Loss.}
Structure loss $L_{st}$ aims to improve separator generation.
It penalizes incorrect separator predictions by: 
$$L_{st} = \sum_{t \in S} l_{t} \cdot missed \cdot w_{miss}$$
where $l_t$ is log probability of token $t$, $S$ separator token locations, $w_{miss}$ miss weight, a hyperparameter, and $missed$ the number of instances where a separator token did not have the highest token probability at its true position.
Structure loss is thus exponential here--for example, if there are 3 missed separators, the loss is
$L=\sum_{n=1}^3{l_t*3*w_{miss}}$, for a total multiplier of 9.

\paragraph{Slot Loss.}
Slot loss is similar to the traditional cross-entropy loss, but reflects the design of SLGM, which provides a set of candidate tokens for generation, whether with a \texttt{<SOURCE>} token or a pre-defined list.
For example, NER predictions should comprise a set of tokens from the source sentence, and one label selected from a pre-defined list of possible labels. 
The loss's key distinction from CE is in the denominator's vocab size, effectively converting a sequence generation task into a classification task.
We implement the final training loss as a weighted sum of cross-entropy, structure, and slot losses, where each coefficient-weight is a hyperparameter.

\subsection{Formatted Decoding during Inference}
\label{sec:formatted-decoding}
During inference, SLGM maintains state information that tracks the current stage of structured text generation within each sentence.
This state reflects which part of the slot is being generated (e.g., head, predicate, or tail) and is implemented as a simple finite state machine.
Each time the model generates a slot separator, the state counter advances, and when a triple separator is generated, the counter resets to zero.
Before producing the next token, SLGM retrieves a format-specific mask based on the current generation state.
This mask restricts the model to generate only legal tokens at that stage.
For example, the loss prevents the model from generating a slot delimiter after a tail or an entity delimiter within an entity, by adding a large negative penalty to the logits of invalid tokens.

\section{Experiments}
\label{sec:Experiments}
Our experiments follow the \textsc{DeepStruct} \citep{deepstruct} experiment protocol, involving two stages: structural pre-training and multi-task training. 
We outline our detailed experimental setup in Appendix \ref{sec:implementation}.

\subsection{Tasks and Datasets}
\paragraph{Structural Pre-training.}
For our structural pre-training, we adapt the TEKGEN training and KELM corpora \citep{tekgen-kelm}, which include NER and RE examples.
However, the corpora do not include information on named entity type, which we address via augmentation by mapping entity types from WikiData \citep{wikidata} to a selected suite of frequent types. 
To supplement the missing type information in TEKGEN and KELM, we map WikiData entities and relations to six coarse-grained entity supertypes—Person, Location, Organization, Product, Terminology, and Event, and manually cluster frequent relation phrases into relation supertypes as described in Appendix \ref{sec:data-stats}. 
The augmentation is discussed in more detail in Appendix \ref{pretrain-mapping}.
In total, we compile 100k sentences from each task and corpus pair for a total of 400k sentences.


\paragraph{Multi-task training.}
After structural pre-training, we perform multi-task training using datasets outlined in Table \ref{tab:Dataset and Format}.
We present dataset statistics in Appendix \ref{sec:data-stats}.
Across 5 tasks: named entity recognition (NER), relation extraction (RE), semantic role labeling (SRL), intent detection (ID), and dialogue state tracking (DST), we employ a total of 13 datasets.
Two datasets are joint entity and relation extraction (JER) datasets, performing NER and RE jointly.
Our dataset formats follow those of TANL \citep{tanl} and \textsc{DeepStruct} \citep{deepstruct}.
This stage lasts for 2 epochs on each dataset.

\begin{table*}
\centering
\begin{tabular}{lll}
\toprule
\textbf{Task} & \textbf{Dataset} & \textbf{Format} \\ \midrule
\multirow{5}{*}{NER} & CoNLL-03 \citep{conll2003} & \multirow{5}{*}{$<$SOURCE$>$ $<$;$>$ instance of $<$;$>$ \textit{tagset} $<$/$>$} \\
& OntoNotes-v5 \citep{ontonotes} & \\
& GENIA \citep{genia} & \\
& CoNLL-04 \citep{conll2004} & \\
& NYT \citep{nyt} & \\ \midrule
\multirow{4}{*}{RE} & TACRED \citep{tacred} & \multirow{4}{*}{$<$SOURCE$>$ $<$;$>$ \textit{tagset} $<$;$>$ $<$SOURCE$>$ $<$/$>$} \\
& TACREV \citep{tacrev} & \\
& CoNLL-04 \citep{conll2004} & \\
& NYT \citep{nyt} & \\ \midrule
SRL & CoNLL-12 \citep{conll2012} & $<$SOURCE$>$ $<$;$>$ instance of $<$;$>$ \textit{tagset} $<$/$>$ \\ \midrule
\multirow{2}{*}{ID} & ATIS \citep{atis} & \multirow{2}{*}{intent $<$;$>$ is $<$;$>$ \textit{tagset} $<$/$>$} \\ 
& SNIPS \citep{snips} & \\ \midrule
DST & MultiWOZ \citep{multiwoz} & [User] $<$;$>$ $<$SOURCE$>$ $<$;$>$ $<$ANY$>$ $<$/$>$ \\ \bottomrule
\end{tabular}
\caption{
Task, dataset and task-specific formats. Joint entity and relation extraction corpora like CoNLL 2004 and NYT dataset split one sentence into two sentences, with different prompt and format.
}
\label{tab:Dataset and Format}
\end{table*}

\begin{table}[ht]
    \centering
    \begin{tabular}{lcccc}
        \toprule
         & SLGM & CE & CE+task & CE+data  \\
         \midrule
         Task name &&&\checkmark&\checkmark \\
         Dataset name &&&&\checkmark \\
         Task instructions & \checkmark&\checkmark&& \\
         Dataset format & \checkmark&\checkmark&&\\
         Format loss & \checkmark &&& \\
         Form. decoding & \checkmark &&& \\
         Fine-tuning & \multicolumn{4}{c}{None; ablated in Section \ref{subsec: fine-tuning}.} \\
         \bottomrule
    \end{tabular}
    \caption{An overview of what features and meta-information SLGM and the 3 baselines use.}
    \label{tab:baselines}
\end{table}

\subsection{Baseline Models}
We establish 3 baseline models across our experiments.
The first, \textbf{CE}, is a Flan-T5-based \citep{flant5} cross-entropy loss model, using SLGM-like input and output format.
The second, \textbf{CE+task}, is the Flan-T5-based cross-entropy loss model, with TANL-style \citep{tanl} prompt with task information.
The third, \textbf{CE+data}, is another Flan-T5-based cross-entropy loss model, but with \textsc{DeepStruct}-style \citep{deepstruct} prompt that includes task and dataset information.
\textsc{DeepStruct} \citep{deepstruct} and TANL \citep{tanl} simply convey task-specific instructions via task and dataset names, as partially illustrated in Figure \ref{fig:intro}.
However, such design relies on the models' ability to retrieve such information from memorization and is likely unreliable and difficult to generalize.
\textbf{SLGM} breaks them down in a way that is easier to access, allows the model to reason over the input rather than retrieving the instructions from memory, as also illustrated in Figure \ref{fig:intro}.
The baseline models thus differ from \textbf{SLGM}, which includes a more transparent task and dataset information as explicit description of output format and tagset, and uses additional format loss and formatted decoding for stable structure generation.
We outline a tabular comparison of SLGM and the three baselines in Table \ref{tab:baselines}.


\subsection{Evaluation Method}
We report the performance and the number of format errors on each dataset.
Performance on all tasks but DST is measured with micro F1 score.
For joint entity and relation extraction, we measure F1 score for entity and relation extraction respectively.
For dialogue state tracking task, we use joint accuracy score to measure performance.

We categorize format errors into three types: length mismatch, source mismatch, and tagset mismatch.
Length mismatch means length of generated tuple does not match with given format.
Source mismatch means the element of output violated \texttt{<SOURCE>} slot token restriction.
When the model generates tokens that don't exist in source sentence, source mismatch count increases.
Finally, a tagset mismatch occurs when the model generates an output that is not defined within the tags specified by the dataset.

\begin{table*}[ht]
\centering
\begin{tabular}{c c c r c r c r c r c}
\toprule
\multirow{2}{*}{\textbf{Task}} &
\multirow{2}{*}{\textbf{Dataset}} 
& \multicolumn{2}{c}{\textbf{SLGM}} 
& \multicolumn{2}{c}{\textbf{CE}} 
& \multicolumn{2}{c}{\textbf{CE+task}} 
& \multicolumn{2}{c}{\textbf{CE+data}}
& \textsc{\textbf{DeepStruct}} \\ 
\cmidrule(lr){3-4} \cmidrule(lr){5-6} \cmidrule(lr){7-8} \cmidrule(lr){9-10} \cmidrule(lr){11-11}
 &  & Score & FE & Score & FE & Score & FE & Score & FE & Score\\ 
\midrule
\multirow{3}{*}{\textbf{NER}} 
& CoNLL-03 & 80.28 & 43 & 69.32 & 4700 & 12.06 & 10273 & 87.11 & 5 & 93.1\\
& OntoNotes & 75.87 & 142 & 75.44 & 766 & 24.79 & 5143 & 81.12 & 348 & 87.6\\
& GENIA & 69.88 & 5 & 67.05 & 147 & 65.26 & 70 & 66.48 & 30 & 80.2\\
\midrule
\multirow{2}{*}{\textbf{RE}} 
& TACREV & 71.39 & 2 & 66.95 & 6 & 66.52 & 21 & 67.04 & 13 & -\\
& TACRED & 63.11 & 2 & 58.41 & 6 & 59.51 & 21 & 59.07 & 13 & 74.9\\
\midrule
\multirow{4}{*}{\textbf{JER}} 
& \multirow{2}{*}{CoNLL-04} & 71.74 & 0 & 0.00 & 873 & 0.00 & 827 & 74.88 & 14 & 88.4 \\
&  & 27.87 & 2 & 13.31 & 413 & 23.08 & 491 & 48.53 & 46 & 72.8\\
& \multirow{2}{*}{NYT} & 88.80 & 149 & 88.68 & 322 & 74.60 & 412 & 88.45 & 23 & 95.4 \\
&  & 59.36 & 20 & 65.80 & 17 & 45.07 & 464 & 67.47 & 7 & 93.7 \\
\midrule
\multirow{1}{*}{\textbf{SRL}} 
& CoNLL-12 & 83.45 & 0 & 82.47 & 161 & 82.37 & 158 & 82.35 & 159 & 60.6\\
\midrule
\multirow{2}{*}{\textbf{ID}} 
& ATIS & 93.96 & 0 & 94.09 & 11 & 94.30 & 15 & 94.21 & 9 & 97.3\\
& SNIPS & 96.86 & 0 & 96.58 & 0 & 95.79 & 1 & 96.07 & 1 & 97.4\\
\midrule
\multirow{1}{*}{\textbf{DST}} 
& MultiWOZ & 38.87 & 0 & 38.16 & 667 & 36.68 & 0 & 37.18 & 0 & 53.5\\
\midrule
\midrule
\multicolumn{2}{c}{\textbf{Average}} & 70.88 & 28.08 & 62.79 & 622.23 & 52.31 & 1376.62 & 73.07 & 51.38 & 82.9\\
\bottomrule
\end{tabular}
\caption{
Main result of our experiment on multi-task setting. FE stands for "Format Error". Higher scores are better, while fewer format errors are better. Average scores and format errors are dataset-wise macro average.
\textsc{DeepStruct} does not report the number of format errors.
}
\label{tab:Main result}
\end{table*}

\section{Results and Discussion}
\label{sec:result}

In Section \ref{sec:plain-results}, we compare our \textbf{SLGM} framework to other baselines.
In Sections  \ref{subsec:formatted-decoding-results} and \ref{sec:fe-results}, we discuss ablation on formatted decoding (FD) and the distribution of format errors with and without FD.
In Sections \ref{subsec: fine-tuning} through \ref{subsec: low resource}, we perform ablation studies on fine-tuning, model size, and low-data scenarios.
Sections \ref{sec:plain-results} through \ref{subsec:formatted-decoding-results} report an average over 5 runs across seeds; Sections \ref{subsec: fine-tuning} and after report results from single runs.
Full tables with SLGM and baseline performance on every dataset in every setting are shown in Appendix \ref{appendix: detailed experiment result}.

\subsection{Main results}
\label{sec:plain-results}

We report our main results in Table \ref{tab:Main result}, comparing our \textbf{SLGM} framework to three baselines: \textbf{CE} without any meta-information, \textbf{CE+task} with task information, and \textbf{CE+data} with task and dataset information.
Training and predicting without dataset information (\textbf{CE}), performance drops by 11 points compared to when dataset information is included (\textbf{CE+data}), and the frequency of format errors increases 10-fold.
\textbf{SLGM} performance excels against baseline models without dataset information, \textbf{CE} and \textbf{CE+task} and is comparable to those of dataset-aware \textbf{CE+data} on most datasets.
Even without explicit dataset information, \textbf{SLGM} is also more reliable than our baseline models, generating fewer format errors than the dataset-aware \textbf{CE+data}.
This suggests \textbf{SLGM} is likely to be robust in out-of-distribution prediction with varying data format or tagset, offering competent performance without even specific dataset information.

At the same time, we acknowledge that \textbf{CE} and \textbf{CE+task} models have no access to any tagset information, unlike \textbf{SLGM}, whose task instructions provide tagset information.
Since there are multiple datasets for NER and JER, each with its unique tagset, format errors are more likely.
On the other hand, RE and ID datasets share identical or similar tagsets, resulting in a low number of format errors.
The same is observed in SRL and DST, with only one dataset each.
In the following subsection, we further analyze the distribution of format errors and experiment to determine whether the use of formatted decoding as a means of providing indirect tagset information is effective for baseline models as well.

\subsection{Ablation: Formatted Decoding at Inference}
\label{subsec:formatted-decoding-results}
To investigate the impact of formatted decoding on generating the correct format, we ablate formatted decoding in our baseline and \textbf{SLGM} models.
Table \ref{tab:formatted-decoding} shows the average score and average format errors across all datasets, with and without formatted decoding during inference.
The numbers suggest that formatted decoding is helpful for extracting the correct tagsets.
In the \textbf{CE} baseline model, despite having been trained without dataset information, formatted decoding led to an increase of 6 points in F1 score and a decrease of 94\% in format errors.

However, when trained with dataset information, the model suffers from the absence of dataset information even with format information, as seen in \textbf{CE+task} performance with and without formatted decoding.
This suggests that the model relies on dataset-specific distributions during training, and removing that signal at inference prevents it from generalizing properly. In this sense, formatted decoding acts as a structural inductive bias, guiding the model to produce correct formats, but it cannot fully compensate for missing dataset cues.
Formatted decoding does increase F1 score and decrease the number of format errors, but the number of format errors is still high, compared to other baseline and \textbf{SLGM} models that use formatted decoding.
From this, we find that when designing models with real-world applications in mind, including dataset names as a structure cue during training can lead to undesirable effects, hindering performance and generalization.

Without format-aware decoding, models trained with the format loss still achieved the highest scores and produced the fewest format errors.
This indicates that, even when implicit, format loss provides a useful signal that helps the model infer the task, dataset type, and output structure. 
The slot losses contribute additional tagset-specific supervision.
Because they are weighted more heavily than cross-entropy, which distributes gradients across all tokens, they exert stronger pressure on the model to learn which spans to extract and how to label them. 
Together, format loss not only guides the models toward a more explicit understanding of the sentence’s structure, but also helps models produce well-formed structured outputs.
We outline the full result on each dataset in Appendix \ref{sec:formatted-decoding-appendix}.

\subsection{Format Error Analysis}
\label{sec:fe-results}

\begin{figure}[t!]
    \centering
    \includegraphics[width=\linewidth]{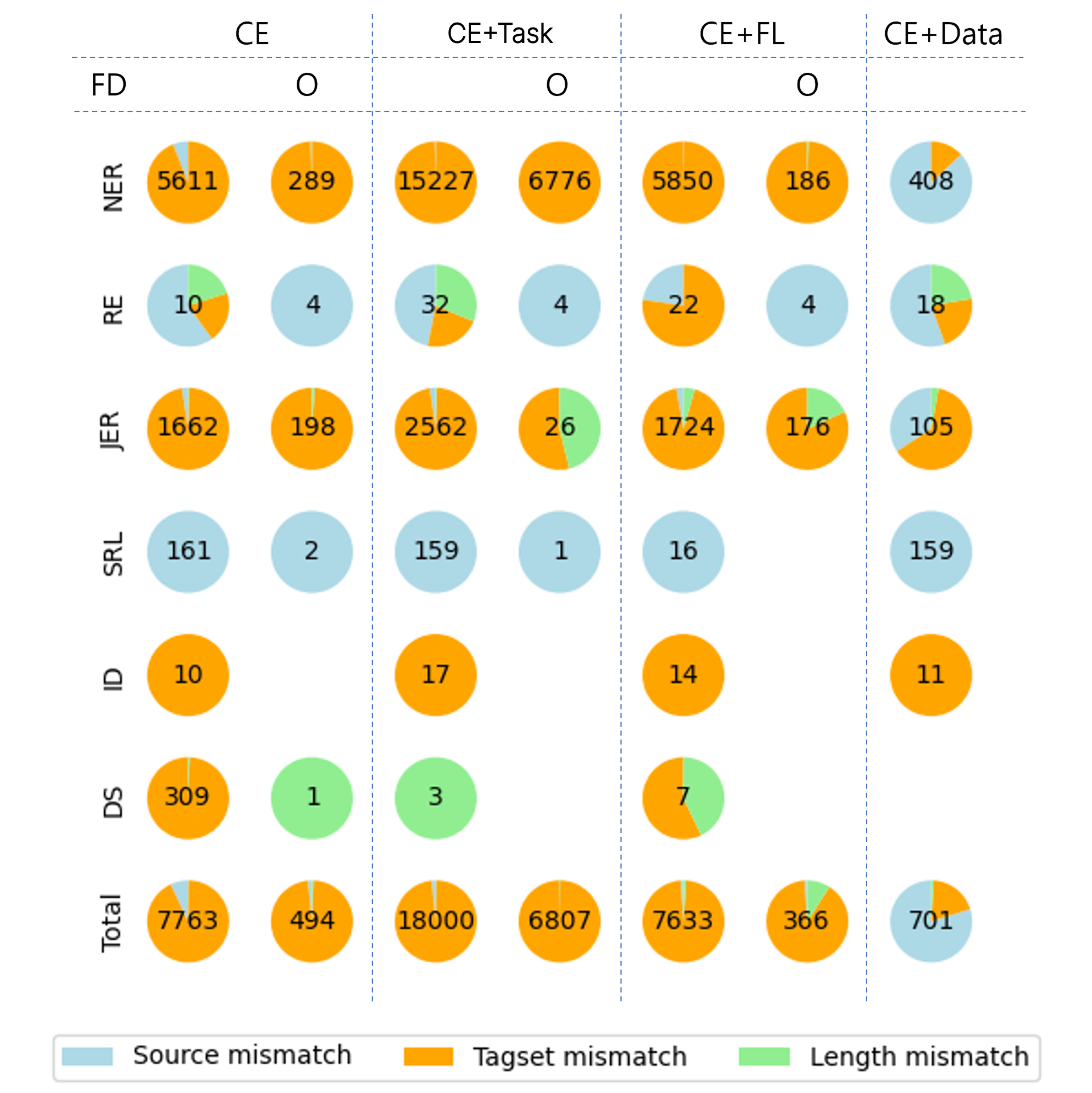}
    \caption{Format error frequency and ratio for each task. Numbers indicate the total number of FEs across every dataset inside given task.}
    \label{fig:format_error_analysis}
\end{figure}

We analyze the frequency of format errors produced by each model, both with and without formatted decoding.
Figure \ref{fig:format_error_analysis} summarizes the total error counts and the distribution of error types.
As shown in the figure, the absence of dataset information in  \textbf{CE+task} most prominently leads to tagset mismatches: without dataset-specific cues, the model struggles to generate well-formed output strings.
Conversely, when dataset information is provided, the model may overfit to the supplied tagsets and therefore exhibits more source prediction errors, as seen in \textbf{CE+data}.
Formatted decoding substantially reduces tagset mismatches, though it does not eliminate them entirely because it constrains only individual tokens rather than full output sequences.
Most remaining mismatches involve producing shorter tags with equivalent meaning, like generating \texttt{LOC} instead of \texttt{LOCATION}.
In \textbf{CE+data}, trained with explicit dataset information, source mismatches are more common. Because this model can reliably identify legal tagsets, its errors primarily reflect difficulties in predicting the correct source tokens, especially in NER.

~

\begin{table}[t!]
\centering
\begin{tabular}{lcr}
\toprule
\textbf{Loss} & \textbf{Avg. F1} & \textbf{Avg. FE} \\
\midrule
CE & 63.03 & 625 \\
\quad + FD & 69.87 & 45 \\
\quad + FL & 63.51 & 582 \\
\quad + FL + FD (=SLGM) & \textbf{70.86} & \textbf{29} \\
\midrule
CE+task & 52.73 & 1405 \\
\quad + FD & 59.32 & 569 \\
\bottomrule
\end{tabular}
\caption{Average score with respect to loss and inference method. 
FD is expressed as an indented row. 
SLGM corresponds to CE + Format Loss + Formatted Decoding.}
\label{tab:formatted-decoding}
\end{table}

\subsection{Ablation Study: Fine-tuning}
\label{subsec: fine-tuning}
\begin{table}[t]
\centering
\begin{tabular}{l c r}
\toprule
\textbf{Loss} & \textbf{Avg. F1} & \textbf{Avg. FE} \\
\midrule
CE & 53.10 & 742 \\
\quad + FT & 77.70 & 56 \\
\quad + FL & 66.74 & 504 \\
\quad + FL + FT & 78.93 & 16 \\
\midrule
CE+task & 53.15 & 1404 \\
\quad + FT & 63.53 & 606 \\
\midrule
\midrule
SLGM& 73.37 & 19 \\
\quad + FT & 78.85 & \textbf{1} \\
\midrule
\textsc{DeepStruct} & 82.9 & - \\
\quad + FT & \textbf{84.9} & -\\
\bottomrule
\end{tabular}
\caption{Average score and format error with and without fine-tuning (+FT).
SLGM corresponds to CE + FL + FD.
\textsc{DeepStruct} \citep{deepstruct} does not report the number of format errors.}
\label{tab:fine-tuning}
\end{table}

We next examine the effects of fine-tuning to SLGM.
Since fine-tuning involves optimizing on a single dataset with clean tagset, we expect it to improve performance even for models without dataset information, reducing the comparative merit of \textbf{SLGM} to baseline models without dataset information \textbf{CE} and \textbf{CE+task}.
Starting from the models trained in the multi-task setting, we fine-tune the models on each dataset for 5 epochs and evaluate their performance. 
The results are shown in Table \ref{tab:fine-tuning}.
Compared to results with formatted decoding shown in Table \ref{tab:formatted-decoding}, fine-tuning generally yields higher F1 scores.
As expected, for \textbf{CE} with format loss and \textbf{SLGM}, formatted decoding contributes only marginal additional gains, suggesting a conceptual connection between formatted decoding and fine-tuning: while fine-tuning explicitly adapts the model to dataset-specific distributions, formatted decoding provides a lightweight mechanism that captures some of the same benefits when task-specific adaptation is not feasible.
This supports our interpretation of formatted decoding as a partial surrogate for fine-tuning—similar in spirit to parameter-efficient adapters \citep{adapter}, which leave the base model unchanged while adding small task-specific components.

\subsection{Ablation Study: Model Size}
\label{subsec: model size}
\begin{figure}[t!]
    \centering
    \includegraphics[width=\linewidth]{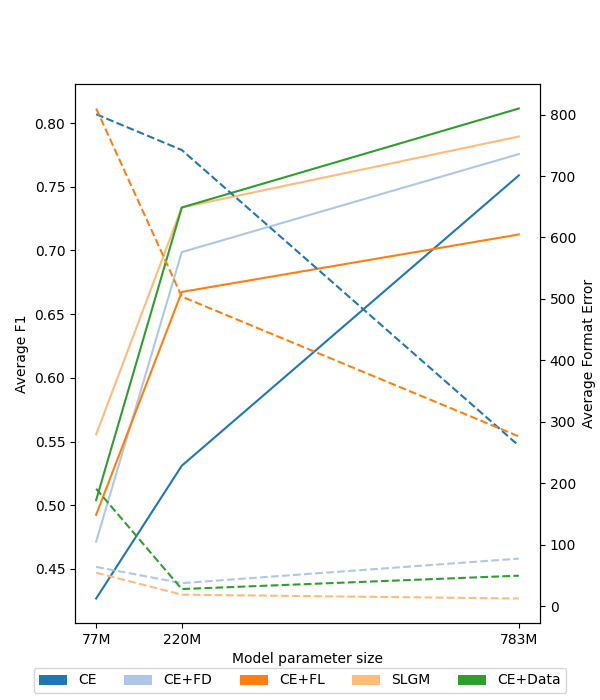}
    \caption{Average score and format error according to model size: \texttt{small}, \texttt{base}, and \texttt{large}. Solid line indicates Average F1 score (left y-axis). Dotted line indicates format errors (right y-axis).}
    \label{fig:model-size-compare}
\end{figure}

We additionally evaluate the SLGM framework with \texttt{Flan-T5-small} (77M params) and \texttt{Flan-T5-large} with (0.8B), with the same hyperparameters as described in Section \ref{sec:Experiments}. We outline a summary of results in Figure \ref{fig:model-size-compare}.

On the \texttt{small} model, our method \textbf{SLGM} even outperforms \textbf{CE+data}, with access to dataset-specific information. The \texttt{small} model with a more limited capacity, appear to benefit from the stronger supervision provided by the format loss and formatted decoding, which offer more direct cues about the correct labels than plain cross-entropy. 
For the \texttt{large} model, a standard cross-entropy objective already yields substantially better results than \texttt{small}: with sufficient capacity, the model is able to infer dataset characteristics and retrieve the correct structure information without explicit cues.
Interestingly, we observe that using the format loss alone with the \texttt{large} model can hurt performance.
Using format loss and formatted decoding together, as done in \textbf{SLGM}, is still effective.

Using our best configuration, we achieve state-of-the-art performance on several benchmarks (Table~\ref{tab:model-size-full}). In particular, we obtain 94.8 micro-F1 on CoNLL-03 \citep{conll2003} NER and 98.3 micro-F1 on ATIS intent detection \citep{atis}.
These findings indicate that our method can be further strengthened through fine-tuning, and that it can deliver strong performance even with smaller $<$1B-parameter models.

\subsection{Ablation Study: Low-Resource Settings}
\label{subsec: low resource}
\begin{figure}[t!]
    \centering
    \includegraphics[width=\linewidth]{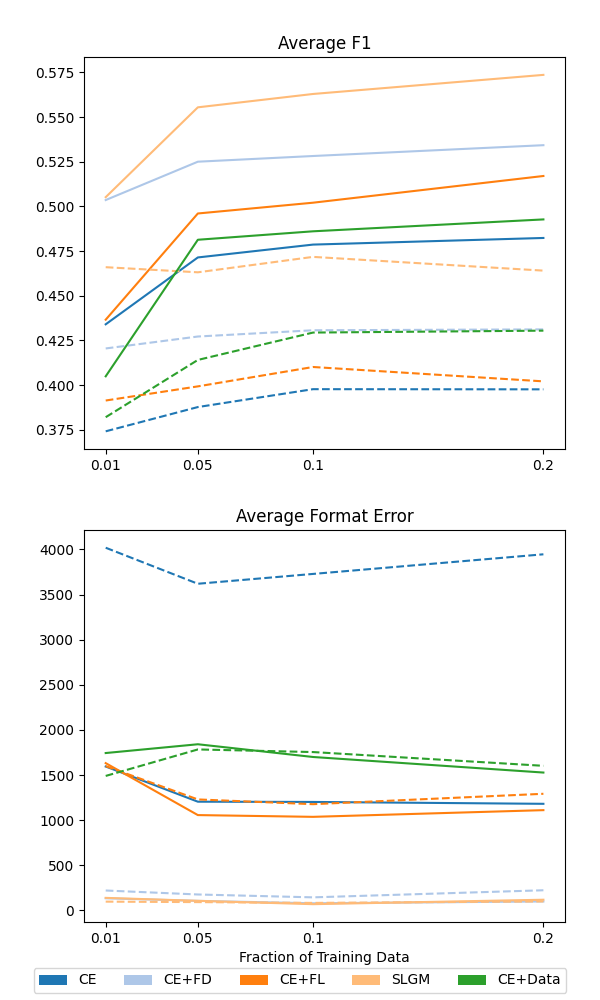}
    \caption{Average score and format errors according to dataset fraction. Solid line represents model with structural pre-training, and dotted line represents model without structural pre-training.}
    \label{fig:low-resource}
\end{figure}


Our analysis indicates that with SLGM, even smaller models are able to effectively encode structure in data and reliably retrieve entities, relations, and other structural information.
To test the robustness of the framework, we also investigate its behavior in low-resource settings in the data side.
We operationalize low-resource settings in two ways: 1) no structure pre-training and 2) training on a subset (1\%, 5\%, 10\%, and 20\%) of available data while preserving label distributions.
We note that our implementation without structure pre-training is equivalent to the traditional NLP pipeline of fine-tuning after pre-training \citep{devlin-etal-2019-bert, linzen-2020-accelerate}.
For each multi-task dataset, we label-proportionally sample training subsets, ensuring at least one instance of each label.
In this ablation, we exclude MultiWOZ \citep{multiwoz} due to ambiguous label boundaries.

Figure~\ref{fig:low-resource} presents the results.
Across all settings, structure pre-training is conducive to performance gains, echoing findings from prior work \citep{deepstruct, min-etal-2025-punctuation}.
The finding indicates it contributes to the model’s robustness and helps guide sentence interpretation when data is limited.

\textbf{SLGM} consistently outperforms other baseline models, with and without structural pre-training.
SLGM reports higher performance and fewer format errors than the dataset-aware \textbf{CE+data} even under severe data constraints.
We also find that, in low-resource scenarios, formatted decoding is generally more reliable than format loss: without structural pre-training, larger training data does not improve \textbf{CE+FL}, \textbf{SLGM}, which rely on format loss.
While performance remains stable across most datasets, we observe a sharp increase in malformed output and degeneration for the NYT relation extraction task \citep{nyt}, which we attribute to sparse supervision \citep{li2023repetition}.

\section{Conclusion}


In this work, we introduce the Structured Language Generation Model, a novel framework for improving formatted structure prediction with generative pre-trained language models via structure-aware input formatting, loss design, and format-aware decoding.
Our experiments on 13 datasets across named entity recognition, relation extraction, semantic role labeling, intent detection, and dialogue state tracking show that SLGM consistently enhances the structure prediction and entity retrieval abilities of $<$1B parameter models. 
SLGM achieves these gains via improved alignment between structure and output, without additional model parameters or dataset-specific engineering, and can act as a zero-weight adapter that approximates task-specific fine-tuning in low-resource settings.
Furthermore, the task- and model-agnostic nature of the framework may allow easy integration with LLM-powered retrieval \citep{lewis2020retrieval} or guided generation systems \citep{willard2023efficientguidedgenerationlarge}, that has been shown to benefit from such alignment \citep{zheng2024sglang, li2025structrag}.
Our results highlight the value of aligning GLMs’ internal structural knowledge with their output space, and suggest a promising direction for building more robust, general-purpose structured prediction and knowledge retrieval systems using generative models.

\section*{Limitations and Future Work}


Format loss and formatted decoding represent basic strategies designed to compel the model to produce outputs in accordance with a predetermined format. There may be many improvements on both scoring and real-world usage.
There may be many problems in specific situations.
Suppose we extract named entity with location but there is no location tag in format. 
The model expected a location and extracted some words, but the location tag is illegal. 
In plain decoding strategy, the model would generate any token similar to location.
However, when using formatted decoding, it would generate unexpected string, which could be worse than extracting a wrong tag.
We did not conduct quantitative experiments regarding this situation.

This scenario could potentially be addressed by employing a beam decoding strategy, which operates on triple units.
This issue is related to the fact that the model does not know about format during actual attention calculation.
Despite the explicit format being given, model cannot utilize format information at attention layers. 
We think additional cross attention layer over formats may show better results. We leave these possible improvements as future work.

Finally, while we predict that the task- and model-agnostic nature of \textbf{SLGM} will allow integration with retrieval-augmented generation (RAG) systems \citep{lewis2020retrieval} and format guided generation \citep[e.g. with regular expressions; ][]{willard2023efficientguidedgenerationlarge}, we do not explicitly show this.
The community may benefit from an explicit investigation into the efficacy of SLGM's components in these areas.

\section*{Responsible Research Statement}
In our research, we utilized a pretrained model and made use of publicly available datasets published on the web.
We acknowledge that the data obtained from the web may contain potential biases. 
Our model can bear potential risks and harms discussed in \citet{gpt3}, and we clarify that our research did not specifically address or consider these risks.
We ensured that all datasets employed in our study were accessed and used in a manner that respects their intended use and complies with any associated licenses or terms of service.
We are also mindful of the potential biases present in these datasets and the pretrained model.

We used ChatGPT's GPT-4 \citep{openai_chatgpt} as a debugging and text refining tool.
We acknowledge and address the ethical considerations associated with the use of such AI technology, and have thoroughly reviewed the content to ensure that it does not include any unethical material.

\section*{Acknowledgements}
This paper is based on our work performed at NC AI.
Our work was partially made possible by the NC AI computing cluster managed by Andrew Matteson at the time and the NC AI Co-op Internship program.
We report the following contribution statement:
Minho Lee designed and implemented the experiments;
Minho Lee, Junghyun Min, and Woochul Lee collected and pre-processed datasets used in this work;
Minho Lee, Junghyun Min, and Yerang Kim performed data analysis, literature review, and participated in writing;
Woochul Lee and Yeonsoo Lee provided and high-level guidance in this work.
We thank Chunghee Lee and anonymous reviewers for their helpful comments in improving this work.

\bibliography{AAAI26/slgm}

\appendix

\section{GPT-4 Instructions}
\label{appendix:chatgpt-instruction}
Among the results shown in Table \ref{tab:deepstruct_dataset_ablation} is GPT-4 \citep{openai_chatgpt} performance. 
We use a custom prompt definining the task and label space, with two examples to show the expected output format. A full example prompt is shown below:

\begin{lstlisting}
You are given multiple sentences. You have to extract relations between entities in given sentences. 
For each sentence, extract every named entity that have relations first. Types of named entity must be one of "location", "organization", "other", "human". Then, extract relations between extracted named entities. Types of relation must be one of "kills", "lives in", "works for", "located in", and "organization based in".

You must FOLLOW given structure. When extracting named entities, you must extract in this form: [ENTITY] <;> instance of <;> [TYPE] . When there are multiple named entitis. separate with </> .  When extracting relations, you must extract in this form: [HEAD] <;> [TYPE] <;> [TAIL] . Same with named entity extraction, if there are multiple relations, separate with </> .

For each sentence, first line should be result of named entity extraction, and second line should be result of relation extraction. When given multiple sentence(separated by empty line), pad empty line between extractions.

When given a file, each line contains single sentence.

Example

Input

John Wilkes Booth , who assassinated President Lincoln , was an actor .

The opera company performed at the Palace of Fine Arts , in San Francisco , on June 30 and July 1-2 , said Kevin O 'Brien , a spokesman for the theater.

<input end>

Output

John Wilkes Booth <;> instance of <;> human </> President Lincoln <;> instance of <;> human

John Wilkes Booth <;> kills <;> President Lincoln

Palace of Fine Arts <;> instance of <;> location </> San Francisco <;> instance of <;> location </> June 30 <;> instance of <;> other </> July 1-2 , <;> instance of <;> other </> Kevin O 'Brien <;> instance of <;> human

Palace of Fine Arts <;> located in <;> San Francisco

<output end>
\end{lstlisting}

With the given prompt, we measure GPT-4 performance on the first 40 examples of CoNLL-04 joint entity recognition and relation classification's test split \citep[JER; ][]{conll2004}.

\section{Details on Dataset Statistics and Mapping to Supertypes}
\label{pretrain-mapping}
\label{sec:data-stats}


\paragraph{Dataset statistics.} Table \ref{tab:dataset-statistics} shows the statistics for each multi-task dataset.
Note that TACRED \citep{tacred} and TACREV \citep{tacrev} share same number of sentences; their difference is on validation and test labels.
When training with format loss, we remove sentences which have labels violating format errors, which accounts for less than 1\% of the data.
On evaluation, we used every sentences as-is, without manual validation.

\begin{table}
\centering
\begin{tabular}{lr}
\toprule
\textbf{Type} & \textbf{Count} \\
\midrule
Location & 10,594,011 \\
Person & 10,239,553 \\
Organization & 3,182,685 \\
Product & 2,233,553 \\
Term & 1,186,434 \\
Event & 514,083 \\
\bottomrule
\end{tabular}
\caption{Manually compiled entity supertypes.}
\label{tab:ent-type-mapping}
\end{table}

\paragraph{Entity supertypes.}
In Section \ref{sec:Experiments}, we describe that entities and relation types have been mapped to supertypes.
Among entity types, we filter out those that appear less than 20k times to remove the long tail.
Then, we manually labeled them into 6 different entity types: Person, Location, Organization, Product, Terminology, and Event. 
Surface forms that we could not find from WikiData are not used. 
In Table \ref{tab:ent-mapping-example}, we illustrate examples of entity mapping for each entity supertype.

\paragraph{Relation supertypes.}
Similarly, we collect relations containing entities that have valid types mapped at Table \ref{tab:ent-mapping-example}. 
Among these relations, we filtered out relations that appear less than 10k times, again to remove the long tail and reduce noise.
There were 168 relation phrases meeting these conditions. 
We additionally clustered 82 of 168 relation phrases into 35 similar clusters.
Table \ref{tab:rel-type-mapping} shows statistics of mapped relations. 

\begin{table}[thbp]
\centering
\begin{tabular}{l l r r r}
\toprule
\textbf{Task} & \textbf{Dataset} & \textbf{Train} & \textbf{Dev} & \textbf{Test} \\
\midrule
\multirow{3}{*}{NER} 
 & CoNLL-03 & 14,986 & 3,465 & 7,148 \\
 & OntoNotes-v5 & 75,187 & 9,603 & 9,479 \\
 & GENIA & 15,023 & 1,669 & 1,854 \\
\midrule
\multirow{2}{*}{JER} 
 & CoNLL-04 & 922 & 231 & 288 \\
 & NYT & 56,196 & 5,000 & 5,000 \\
\midrule
RE & TACRED* & 68,124 & 22,631 & 15,509 \\
SRL & CoNLL-12 & 253,070 & 35,297 & 24,462 \\
\midrule
\multirow{2}{*}{ID} 
 & ATIS & 4,478 & 500 & 893 \\
 & SNIPS & 13,084 & 700 & 700 \\
\midrule
DST & MultiWOZ & 62,367 & 7,371 & 7,368 \\
\bottomrule
\end{tabular}
\caption{Number of examples in each multi-task dataset. TACRED \citep{tacred} and TACREV \citep{tacrev} contain the same amount of data, but vary in their dev and test splits.}
\label{tab:dataset-statistics}
\end{table}

\begin{table}[t!]
\centering

\begin{tabular}{lrlr}
\toprule
\textbf{Type} & \textbf{Rel count} & \textbf{Type} & \textbf{Rel ount} \\
\midrule
country & 3,233,250 & member of & 1,768,459 \\
located in & 1,657,574 & product of & 1,034,479 \\
date of birth & 1,009,859 & occupation & 937,797 \\
instance of & 879,955 & place of birth & 737,735 \\
educated at & 496,122 & cast member & 446,835 \\
works at & 353,896 & date of death & 300,792 \\
awarded & 299,818 & contains & 253,338 \\
distributor & 162,945 & part of & 146,352 \\
gender & 142,878 & place of death & 122,463 \\
start date & 110,483 & sibling & 92,805 \\
child & 84,473 & parent & 81,600 \\
owner & 77,977 & genre & 76,344 \\
spouse & 76,093 & nominated & 69,833 \\
winner & 61,688 & position & 60,142 \\
created & 49,232 & background & 42,186 \\
capital & 39,123 & twin towns & 38,671 \\
capital of & 31,669 & president & 20,402 \\
family & 15,840 & & \\
\bottomrule
\end{tabular}
\caption{Manually compiled relation supertypes. For example, relations `country', `sovereign state', `historical country' are mapped to `country'.
}
\label{tab:rel-type-mapping}
\end{table}

\begin{table}[ht]
\centering
\begin{tabular}{lcll}
\toprule
\textbf{Original type} & \textbf{Count} & \textbf{Map} & \textbf{Example} \\
\midrule
Human & 9.7M & Person & \multirow{2}{5em}{Umberto II of Italy} \\\\
Country & 4.2M & Loc. & England \\
Taxon & 1.2M & -- & \multirow{2}{5em}{Natuna Island surili} \\\\
\multirow{2}{7em}{Association football club} & 954k & Org. & FC Baden \\ \\
Film & 515k & Product & Avengers \\
\multirow{2}{7em}{Summer Olympic Games} & 185k & Event & \multirow{2}{6em}{2024 Summer Olympics} \\\\
\bottomrule
\end{tabular}
\caption{Entity type mapping examples.}
\label{tab:ent-mapping-example}
\end{table}

\section{Implementation Detail}
\label{sec:implementation}
We use the Flan-T5~\citep{flant5}, a family of instruction-tuned models based on T5 \citep{t5} from HuggingFace hub \citep{huggingface} as our base model.
As shown in Figure \ref{fig:intro}, we used "Extract \textit{target} from: \textit{sentence}" as our input prompt, where \textit{target} is determined by task. For \textbf{CE+task} and \textbf{CE+data} models, we add additional format information between task and sentence.
Our models are based on the base sized checkpoint with 220M parameters, trained with batch size 16 per GPU on 4 NVIDIA A100 GPUs with 40G memory. 
For SLGM hyperparameters, we used $w_{ce}=$ 0.5, $w_{st}=$ 0.2, $w_{miss}=$ 0.33, and $w_{sl}=$ 0.3. 
We used greedy decoding when generating answers.

\section{Detailed experiment result}
\label{appendix: detailed experiment result}
This section shows full result of Section \ref{sec:result}. We report scores and format errors on every dataset for each settings.

\subsection{Formatted decoding}
\label{sec:formatted-decoding-appendix}

\begin{table*}[t]
\setlength{\tabcolsep}{2pt}
\centering
\begin{tabular}{|c|c|c|c|c|c|c|c|c|c|c|c|c|c|}
\hline
\multirow{2}{*}{\textbf{Task}} & \multirow{2}{*}{\textbf{Dataset}} & \multicolumn{2}{c|}{\textbf{CE}} & \multicolumn{2}{c|}{\textbf{CE+FD}} & \multicolumn{2}{c|}{\textbf{CE+FL}} & \multicolumn{2}{c|}{\textbf{SLGM}} & \multicolumn{2}{c|}{\textbf{CE+task+FD}} & \multicolumn{2}{c|}{\textbf{CE+data+FD}} \\ \cline{3-14}
 &  & Score & FE & Score & FE & Score & FE & Score & FE & Score & FE & Score & FE \\ \hline
\multirow{3}{*}{\textbf{NER}} & CoNLL-03 & 69.32 & 4700 & 78.09 & 79 & 66.17 & 5127 & 80.28 & 43 & 28.30 & 2196 & 87.11 & 0 \\
 & OntoNotes & 75.44 & 766 & 76.04 & 288 & 75.50 & 572 & 75.87 & 142 & 26.72 & 5164 & 81.64 & 45 \\
 & GENIA & 67.05 & 147 & 66.31 & 9 & 70.31 & 75 & 69.88 & 5 & 64.95 & 3 & 66.35 & 1 \\
\hline
\multirow{2}{*}{\textbf{RE}} & TACREV & 66.95 & 6 & 66.93 & 2 & 71.51 & 12 & 71.39 & 2 & 66.48 & 2 & 67.00 & 2 \\
 & TACRED & 58.41 & 6 & 58.42 & 2 & 63.21 & 12 & 63.11 & 2 & 59.48 & 2 & 59.05 & 2 \\
\hline
\multirow{4}{*}{\textbf{JER}} & \multirow{2}{*}{CoNLL-04} & 0.00 & 873 & 66.73 & 0 & 0.00 & 873 & 71.74 & 0 & 66.02 & 0 & 74.50 & 0 \\
 &  & 13.31 & 413 & 26.49 & 3 & 16.84 & 493 & 27.87 & 2 & 27.23 & 2 & 45.71 & 2 \\
 & \multirow{2}{*}{NYT} & 88.68 & 322 & 88.45 & 156 & 88.84 & 300 & 88.80 & 149 & 73.79 & 17 & 88.36 & 1 \\
 &  & 65.80 & 17 & 65.75 & 0 & 59.42 & 33 & 59.36 & 20 & 45.03 & 8 & 67.47 & 2 \\
\hline
\multirow{1}{*}{\textbf{SRL}} & CoNLL-12 & 82.47 & 161 & 82.35 & 3 & 83.44 & 19 & 83.45 & 0 & 82.27 & 1 & 82.24 & 1 \\
\hline
\multirow{2}{*}{\textbf{ID}} & ATIS & 94.09 & 11 & 93.96 & 0 & 94.14 & 12 & 93.96 & 0 & 93.85 & 0 & 94.07 & 0 \\
 & SNIPS & 96.58 & 0 & 96.58 & 0 & 96.86 & 0 & 96.86 & 0 & 95.72 & 0 & 96.01 & 0 \\
\hline
\multirow{1}{*}{\textbf{DST}} & MultiWOZ & 38.16 & 667 & 38.21 & 0 & 38.85 & 7 & 38.87 & 0 & 36.68 & 0 & 37.14 & 0 \\
\hline
\hline
\multicolumn{2}{|c|}{\textbf{Average}} & 62.79 & 622.23 & 69.56 & 41.69 & 63.47 & 579.62 & 70.88 & 28.08 & 58.96 & 568.85 & 72.82 & 4.31 \\
\hline
\end{tabular}
\caption{Full result of F1 scores and format errors regarding to formatted decoding.}
\label{tab:formatted-decoding-full}
\end{table*}

Table \ref{tab:formatted-decoding-full} shows the full result using formatted decoding (Section \ref{subsec:formatted-decoding-results}), averaged on 5 runs. Comparing with Table \ref{tab:Main result}, formatted decoding greatly reduces format errors on NER and JER tasks. It also shows great score improvements on low resource data like CoNLL-04.
Even if the model knows about dataset information, formatted decoding still upgrades score and reduces format errors.

\subsection{Fine-tuning}
\label{sec:fine-tuning-appendix}
\begin{table*}[t]
\setlength{\tabcolsep}{2pt}
\centering
\begin{tabular}{|c|c|c|c|c|c|c|c|c|c|c|c|c|c|}
\hline
\multirow{2}{*}{\textbf{Task}} & \multirow{2}{*}{\textbf{Dataset}} & \multicolumn{2}{c|}{\textbf{CE+FT}} & \multicolumn{2}{c|}{\textbf{CE+FD+FT}} & \multicolumn{2}{c|}{\textbf{CE+FL+FT}} & \multicolumn{2}{c|}{\textbf{SLGM+FT}} & \multicolumn{2}{c|}{\textbf{CE+task+FT}} & \multicolumn{2}{c|}{\textbf{CE+data+FT}} \\ \cline{3-14}     
 &  & Score & FE & Score & FE & Score & FE & Score & FE & Score & FE & Score & FE \\ \hline
\multirow{3}{*}{\textbf{NER}} & CoNLL-03 & 91.63 & 10 & 92.29 & 0 & 93.65 & 4 & 93.65 & 0 & 77.99 & 2733 & 91.65 & 7 \\
 & OntoNotes & 82.26 & 376 & 82.23 & 4 & 84.22 & 11 & 84.21 & 0 & 41.93 & 2860 & 82.29 & 365 \\
 & GENIA & 75.78 & 5 & 68.09 & 0 & 75.44 & 6 & 75.42 & 0 & 66.72 & 18 & 75.70 & 4 \\
\hline
\multirow{2}{*}{\textbf{RE}} & TACREV & 75.18 & 8 & 74.57 & 3 & 76.95 & 6 & 76.90 & 2 & 73.12 & 7 & 74.58 & 10 \\
 & TACRED & 65.78 & 8 & 64.94 & 3 & 66.87 & 6 & 66.82 & 2 & 64.96 & 7 & 65.51 & 10 \\
\hline
\multirow{4}{*}{\textbf{JER}} & \multirow{2}{*}{CoNLL-04} & 83.50 & 2 & 82.68 & 0 & 85.60 & 3 & 85.48 & 0 & 0.00 & 877 & 84.86 & 1 \\
 &  & 52.43 & 6 & 50.85 & 0 & 55.71 & 14 & 55.24 & 0 & 36.35 & 126 & 50.36 & 4 \\
 & \multirow{2}{*}{NYT} & 90.11 & 11 & 90.08 & 0 & 90.66 & 13 & 90.68 & 0 & 85.28 & 15 & 90.36 & 20 \\
 &  & 74.70 & 35 & 74.47 & 0 & 76.78 & 102 & 76.46 & 3 & 62.83 & 93 & 75.15 & 11 \\
\hline
\multirow{1}{*}{\textbf{SRL}} & CoNLL-12 & 82.75 & 158 & 82.11 & 1 & 84.44 & 11 & 84.45 & 0 & 82.21 & 17 & 82.99 & 163 \\
\hline
\multirow{2}{*}{\textbf{ID}} & ATIS & 97.98 & 2 & 98.09 & 0 & 98.04 & 1 & 97.98 & 0 & 97.70 & 1 & 97.48 & 1 \\
 & SNIPS & 98.43 & 0 & 98.43 & 0 & 96.72 & 0 & 96.72 & 0 & 98.00 & 0 & 98.15 & 0 \\
\hline
\multirow{1}{*}{\textbf{DST}} & MultiWOZ & 39.61 & 0 & 39.35 & 0 & 41.02 & 0 & 41.02 & 0 & 38.85 & 518 & 39.86 & 4 \\
\hline
\hline
\multicolumn{2}{|c|}{\textbf{Average}} & 77.70 & 47.77 & 76.78 & 0.85 & 78.93 & 13.62 & 78.85 & 0.54 & 63.53 & 559.38 & 77.61 & 46.15 \\
\hline
\end{tabular}
\caption{Full result of F1 scores and format errors regarding to dataset specific fine-tuning.}
\label{tab:fine-tuning-full}
\end{table*}

Table \ref{tab:fine-tuning-full} shows full result of fine-tuning (Section \ref{subsec: fine-tuning}) on single run.
Comparing with Table \ref{tab:formatted-decoding-full}, fine-tuning shows 5 to 7 points better F1 score than formatted decoding.
Surprisingly, best setting utilizing fine-tuning is using only format loss, which is better than using both formatted decoding.
Model with format loss win over model with dataset information on almost every dataset.
This is evidence that our format loss can reduce problem as easy as classification.
However, using both fine-tuning and formatted decoding seems making conflict to each other.
This is because some gold instances violates format restriction (e.g. \texttt{<SOURCE>} restriction in named entity recognition).
Without formatted decoding, model can infer tokens that did not appeared in source sentence, when training example exists.
However, we think this is not good for real-world usage, because it can be seen as overfitting on erroneous training examples.

\subsection{Model size}
\label{sec:model-size-appendix}
\begin{sidewaystable*}
\centering
\setlength{\tabcolsep}{2pt}
\begin{tabular}{|c|c|c|c|c|c|c|c|c|c|c|c|c|c|c|c|c|c|c|c|}
\hline
\multirow{2}{*}{\textbf{Task}} & \multirow{2}{*}{\textbf{Dataset}} & \multicolumn{3}{c|}{\textbf{CE}} & \multicolumn{3}{c|}{\textbf{CE+FD}} & \multicolumn{3}{c|}{\textbf{CE+FL}} & \multicolumn{3}{c|}{\textbf{SLGM}} & \multicolumn{3}{c|}{\textbf{CE+data}} & \multicolumn{3}{c|}{\textbf{SLGM+FT}} \\ \cline{3-20}
 &  & S & B & L & S & B & L & S & B & L & S & B & L & S & B & L & S & B & L \\ \hline
\multirow{6}{*}{\textbf{NER}} & \multirow{2}{*}{CoNLL-03} & 48.13 & 64.03 & 85.69 & 58.69 & 80.36 & 86.37 & 54.70 & 72.76 & 83.75 & 70.04 & 84.18 & 89.72 & 71.69 & 89.79 & 92.72 & 88.17 & 93.65 & \textbf{94.80} \\
 &  & 5684 & 5843 & 1756 & 9 & 43 & 5 & 6108 & 4437 & 2531 & 36 & 25 & 18 & 374 & 15 & 45 & 0 & 0 & 0 \\
\cdashline{2-20}
 & \multirow{2}{*}{OntoNotes} & 48.80 & 71.89 & 84.46 & 49.46 & 78.20 & 86.36 & 61.89 & 81.30 & 85.02 & 61.73 & 81.64 & 85.33 & 66.95 & 83.84 & 87.28 & 76.87 & 84.21 & 87.89 \\
 &  & 2080 & 717 & 555 & 360 & 191 & 42 & 1236 & 411 & 291 & 139 & 99 & 81 & 98 & 83 & 375 & 1 & 0 & 0 \\
\cdashline{2-20}
 & \multirow{2}{*}{GENIA} & 45.97 & 0.00 & 77.13 & 41.75 & 66.19 & 72.48 & 50.74 & 72.37 & 76.00 & 50.22 & 72.24 & 75.89 & 46.18 & 67.37 & 76.99 & 64.08 & 75.42 & 76.99 \\
 &  & 193 & 1137 & 5 & 8 & 15 & 7 & 197 & 31 & 16 & 12 & 7 & 4 & 92 & 35 & 0 & 1 & 0 & 0 \\
\cdashline{2-20}
\hline
\multirow{4}{*}{\textbf{RE}} & \multirow{2}{*}{TACREV} & 12.05 & 54.37 & 79.24 & 9.30 & 65.21 & 77.74 & 41.47 & 72.97 & 79.34 & 41.49 & 72.93 & 79.35 & 13.15 & 66.72 & 78.57 & 62.22 & 76.90 & 79.10 \\
 &  & 7 & 5 & 9 & 8 & 2 & 2 & 56 & 7 & 9 & 4 & 2 & 4 & 4 & 7 & 14 & 4 & 2 & 2 \\
\cdashline{2-20}
 & \multirow{2}{*}{TACRED} & 10.72 & 49.18 & 69.34 & 8.24 & 56.92 & 67.53 & 36.81 & 64.35 & 69.23 & 36.80 & 64.32 & 69.25 & 10.86 & 57.70 & 68.73 & 55.35 & 66.82 & 69.24 \\
 &  & 7 & 5 & 9 & 8 & 2 & 2 & 56 & 7 & 9 & 4 & 2 & 4 & 7 & 7 & 15 & 4 & 2 & 2 \\
\cdashline{2-20}
\hline
\multirow{8}{*}{\textbf{JER}} & \multirow{4}{*}{CoNLL-04} & 0.00 & 0.00 & 40.31 & 49.30 & 67.57 & 75.22 & 0.00 & 6.88 & 40.74 & 57.55 & 71.17 & 77.39 & 48.24 & 74.43 & 85.56 & 77.16 & 85.48 & 88.18\\
 &  & 690 & 794 & 643 & 0 & 0 & 0 & 702 & 874 & 635 & 0 & 0 & 0 & 420 & 11 & 1 & 0 & 0 & 0\\
 &  & 0.00 & 4.87 & 45.77 & 7.98 & 27.28 & 45.07 & 0.00 & 23.32 & 38.61 & 9.32 & 33.90 & 44.84 & 0.00 & 46.43 & 59.69 & 26.00 & 55.24 & 60.02\\
 &  & 470 & 530 & 151 & 91 & 0 & 7 & 883 & 425 & 208 & 244 & 2 & 2 & 981 & 99 & 19 & 1 & 0 & 1\\
\cdashline{2-20}
 & \multirow{4}{*}{NYT} & 80.06 & 85.35 & 93.59 & 79.68 & 88.26 & 88.36 & 81.24 & 89.79 & 93.66 & 81.25 & 89.80 & 93.69 & 82.78 & 88.83 & 93.76 & 85.10 & 90.68 & 93.85\\
 &  & 564 & 475 & 93 & 322 & 232 & 939 & 446 & 215 & 77 & 200 & 95 & 47 & 16 & 37 & 0 & 0 & 0 & 0\\
 &  & 45.68 & 54.79 & 84.66 & 45.20 & 66.74 & 84.98 & 29.99 & 65.01 & 85.22 & 29.63 & 64.75 & 85.21 & 49.72 & 66.01 & 86.06 & 38.95 & 76.46 & 86.66\\
 &  & 298 & 82 & 3 & 5 & 1 & 2 & 600 & 120 & 8 & 59 & 11 & 1 & 68 & 38 & 5 & 45 & 3 & 0\\
\cdashline{2-20}
\hline
\multirow{2}{*}{\textbf{SRL}} & \multirow{2}{*}{CoNLL-12} & 65.54 & 80.22 & 87.30 & 65.07 & 82.51 & 86.93 & 72.80 & 84.72 & 87.67 & 72.81 & 84.72 & 87.67 & 65.95 & 82.15 & 87.42 & 73.72 & 84.45 & 87.12 \\
 &  & 167 & 20 & 161 & 2 & 1 & 0 & 83 & 14 & 9 & 4 & 0 & 1 & 164 & 16 & 162 & 4 & 0 & 1 \\
\cdashline{2-20}
\hline
\multirow{4}{*}{\textbf{ID}} & \multirow{2}{*}{ATIS} & 79.81 & 92.64 & 97.87 & 81.38 & 93.72 & 97.87 & 85.25 & 96.51 & 97.92 & 85.78 & 96.52 & 97.76 & 83.92 & 94.36 & 97.92 & 96.19 & 97.98 & \textbf{98.32} \\
 &  & 134 & 31 & 4 & 17 & 0 & 0 & 123 & 8 & 3 & 5 & 0 & 0 & 96 & 10 & 5 & 0 & 0 & 0 \\
\cdashline{2-20}
 & \multirow{2}{*}{SNIPS} & 89.68 & 95.22 & 97.50 & 89.87 & 96.43 & 97.43 & 92.86 & 96.72 & 97.43 & 93.01 & 96.72 & 97.43 & 88.24 & 96.86 & 97.43 & 96.29 & 96.72 & 98.15 \\
 &  & 6 & 1 & 1 & 0 & 0 & 0 & 2 & 0 & 0 & 0 & 0 & 0 & 8 & 0 & 0 & 0 & 0 & 0 \\
\cdashline{2-20}
\hline
\multirow{2}{*}{\textbf{DST}} & \multirow{2}{*}{MultiWOZ} & 28.42 & 37.75 & 43.72 & 26.88 & 38.89 & 42.00 & 32.53 & 40.96 & 42.72 & 32.58 & 40.96 & 42.72 & 27.56 & 39.33 & 42.69 & 35.08 & 41.02 & 42.23 \\
 &  & 106 & 10 & 4 & 0 & 0 & 0 & 36 & 0 & 0 & 0 & 0 & 0 & 155 & 4 & 4 & 0 & 0 & 4 \\
\cdashline{2-20}
\hline
\hline
\multicolumn{2}{|c|}{\multirow{2}{*}{\textbf{Average}}} & 42.68 & 53.10 & 75.89 & 47.14 & 69.87 & 77.57 & 49.25 & 66.74 & 75.18 & 55.56 & 73.37 & 78.94 & 50.40 & 73.37 & 81.14 & 67.32 & 78.85 & 81.73 \\
\multicolumn{2}{|c|}{} & 800.46 & 742.31 & 261.08 & 63.85 & 37.46 & 77.38 & 809.85 & 503.77 & 292.00 & 54.38 & 18.69 & 12.46 & 191.00 & 27.85 & 49.62 & 4.62 & 0.54 & 0.77 \\
\hline
\end{tabular}
\caption{Full result regarding to model size. S, B, L stands for small, base, large, respectively. For each dataset, the upper row is F1 score, and the lower row is format errors. Text with bold means state of the art performance.}
\label{tab:model-size-full}
\end{sidewaystable*}

Table \ref{tab:model-size-full} shows full result regarding to model size (Section \ref{subsec: model size}) on single run. 
On cross-entropy model, model can distinguish characteristics of dataset when model parameters increase.
The power of formatted inference on score is reduced when parameter size increased, yet still effective reducing format errors.
Comparing CE+FL with CE, using only format loss showed worse score and more format error on large sized model.
However, when mixed with formatted decoding, it showed additional increase comparing with models using single methods.
Additionally, when we use every method we tried in this paper (i.e. format loss, formatted decoding, using bigger model, fine-tuning on dataset), we achieved state of the art performance on CoNLL-03 named entity recognition task and ATIS intent detection task.
This means our framework still has room for improvement.

\subsection{Low resource}
\label{sec:low-resource-appendix}

\begin{sidewaystable*}
\centering
\setlength{\tabcolsep}{2pt}
\begin{tabular}{|c|c|c|c|c|c|c|c|c|c|c|c|c|c|}
\hline
\multirow{2}{*}{\textbf{Task}} & \multirow{2}{*}{\textbf{Dataset}} & \multicolumn{4}{c|}{\textbf{CE}} & \multicolumn{4}{c|}{\textbf{CE+FD}} & \multicolumn{4}{c|}{\textbf{CE+FL}} \\ \cline{3-14}
 &  & 0.01 & 0.05 & 0.1 & 0.2 & 0.01 & 0.05 & 0.1 & 0.2 & 0.01 & 0.05 & 0.1 & 0.2 \\ \hline
\multirow{6}{*}{\textbf{NER}} & \multirow{2}{*}{CoNLL-03} & 26.19 & 42.80 & 42.71 & 45.30 & 51.65 & 55.67 & 54.10 & 55.24 & 24.72 & 41.60 & 44.79 & 44.32 \\
 &  & 10169 & 7812 & 7820 & 7853 & 2 & 33 & 32 & 30 & 9937 & 7516 & 7820 & 7822 \\
\cdashline{2-14}
 & \multirow{2}{*}{OntoNotes} & 32.09 & 36.05 & 36.42 & 37.78 & 29.55 & 34.76 & 35.25 & 35.66 & 39.97 & 40.62 & 40.25 & 44.26 \\
 &  & 3599 & 3442 & 3452 & 3392 & 480 & 752 & 612 & 586 & 4705 & 3628 & 3402 & 3472 \\
\cdashline{2-14}
 & \multirow{2}{*}{GENIA} & 42.69 & 45.87 & 45.62 & 42.71 & 38.07 & 39.58 & 39.54 & 38.24 & 45.61 & 44.59 & 46.95 & 46.48 \\
 &  & 379 & 319 & 236 & 217 & 82 & 51 & 43 & 89 & 399 & 290 & 301 & 236 \\
\cdashline{2-14}
\hline
\multirow{4}{*}{\textbf{RE}} & \multirow{2}{*}{TACREV} & 0.06 & 0.00 & 0.00 & 0.00 & 1.08 & 2.98 & 0.89 & 1.69 & 1.70 & 6.99 & 1.52 & 3.02 \\
 &  & 30361 & 29257 & 30610 & 33136 & 1144 & 914 & 623 & 1264 & 39 & 17 & 15 & 20 \\
\cdashline{2-14}
 & \multirow{2}{*}{TACRED} & 7.87 & 5.39 & 10.67 & 7.10 & 3.68 & 2.78 & 1.31 & 0.42 & 1.48 & 6.07 & 1.43 & 2.78 \\
 &  & 75 & 12 & 16 & 16 & 3 & 3 & 3 & 2 & 39 & 17 & 15 & 20 \\
\cdashline{2-14}
\hline
\multirow{8}{*}{\textbf{JER}} & \multirow{4}{*}{CoNLL-04} & 0.00 & 0.00 & 0.00 & 0.00 & 39.00 & 41.08 & 41.64 & 41.35 & 0.00 & 0.00 & 0.00 & 0.00\\
 &  & 762 & 714 & 716 & 740 & 0 & 0 & 1 & 1 & 782 & 702 & 711 & 704\\
 &  & 0.00 & 0.00 & 0.00 & 0.00 & 0.44 & 1.34 & 1.92 & 1.85 & 0.00 & 0.00 & 0.00 & 0.00\\
 &  & 464 & 701 & 691 & 665 & 151 & 42 & 27 & 30 & 446 & 846 & 602 & 943\\
\cdashline{2-14}
 & \multirow{4}{*}{NYT} & 77.64 & 80.34 & 80.66 & 78.57 & 77.22 & 80.24 & 80.32 & 78.35 & 79.33 & 81.23 & 80.92 & 80.56\\
 &  & 1143 & 341 & 459 & 788 & 739 & 244 & 363 & 622 & 1101 & 348 & 493 & 541\\
 &  & 48.18 & 31.60 & 33.65 & 30.24 & 47.83 & 31.15 & 32.04 & 28.94 & 52.19 & 25.60 & 41.50 & 20.19\\
 &  & 134 & 280 & 354 & 292 & 6 & 60 & 22 & 42 & 154 & 572 & 407 & 1482\\
\cdashline{2-14}
\hline
\multirow{2}{*}{\textbf{SRL}} & \multirow{2}{*}{CoNLL-12} & 49.40 & 56.52 & 60.59 & 65.31 & 48.36 & 56.17 & 60.26 & 64.83 & 54.16 & 60.45 & 64.43 & 68.33 \\
 &  & 960 & 413 & 232 & 121 & 30 & 12 & 3 & 4 & 1393 & 680 & 190 & 120 \\
\cdashline{2-14}
\hline
\multirow{4}{*}{\textbf{ID}} & \multirow{2}{*}{ATIS} & 79.54 & 80.07 & 79.45 & 80.64 & 81.20 & 79.48 & 80.99 & 80.59 & 82.70 & 82.68 & 81.39 & 82.29 \\
 &  & 121 & 117 & 129 & 109 & 13 & 10 & 11 & 12 & 142 & 115 & 145 & 124 \\
\cdashline{2-14}
 & \multirow{2}{*}{SNIPS} & 85.37 & 86.69 & 87.56 & 89.57 & 86.59 & 87.45 & 88.59 & 90.30 & 87.84 & 89.37 & 89.02 & 90.34 \\
 &  & 57 & 27 & 27 & 22 & 0 & 0 & 0 & 0 & 47 & 28 & 27 & 25 \\
\cdashline{2-14}
\hline
\multicolumn{2}{|c|}{\multirow{2}{*}{\textbf{Average}}} & 37.42 & 38.78 & 39.78 & 39.77 & 42.05 & 42.72 & 43.07 & 43.12 & 39.14 & 39.93 & 41.02 & 40.21 \\
\multicolumn{2}{|c|}{} & 4018.67 & 3619.58 & 3728.50 & 3945.92 & 220.83 & 176.75 & 145.00 & 223.50 & 1598.67 & 1229.92 & 1177.33 & 1292.42 \\
\hline
\end{tabular}
\caption{Full result of low resource experiment without structural pre-training (Part 1): CE, CE+FD, and CE+FL. For each dataset, upper row is F1 score, and lower row is format errors.}
\label{tab:low-resource-without-pt-full-part1}
\end{sidewaystable*}

\begin{sidewaystable*}
\centering
\setlength{\tabcolsep}{2pt}
\begin{tabular}{|c|c|c|c|c|c|c|c|c|c|}
\hline
\multirow{2}{*}{\textbf{Task}} & \multirow{2}{*}{\textbf{Dataset}} & \multicolumn{4}{c|}{\textbf{SLGM}} & \multicolumn{4}{c|}{\textbf{CE+data}} \\ \cline{3-10}
 &  & 0.01 & 0.05 & 0.1 & 0.2 & 0.01 & 0.05 & 0.1 & 0.2 \\ \hline
\multirow{6}{*}{\textbf{NER}} & \multirow{2}{*}{CoNLL-03} & 55.63 & 60.61 & 59.42 & 61.32 & 26.19 & 44.20 & 45.41 & 45.42 \\
 &  & 14 & 12 & 15 & 20 & 10169 & 9828 & 9686 & 8658 \\
\cdashline{2-10}
 & \multirow{2}{*}{OntoNotes} & 36.00 & 38.80 & 38.78 & 41.39 & 32.09 & 42.05 & 39.90 & 43.55 \\
 &  & 610 & 569 & 592 & 405 & 3599 & 7886 & 8326 & 7794 \\
\cdashline{2-10}
 & \multirow{2}{*}{GENIA} & 44.54 & 43.99 & 45.86 & 46.01 & 42.69 & 48.26 & 47.79 & 47.00 \\
 &  & 57 & 48 & 27 & 68 & 379 & 858 & 879 & 586 \\
\cdashline{2-10}
\hline
\multirow{4}{*}{\textbf{RE}} & \multirow{2}{*}{TACREV} & 1.76 & 7.04 & 1.52 & 3.02 & 11.49 & 4.70 & 12.18 & 9.75 \\
 &  & 4 & 2 & 3 & 3 & 80 & 12 & 19 & 25 \\
\cdashline{2-10}
 & \multirow{2}{*}{TACRED} & 1.53 & 6.12 & 1.43 & 2.78 & 5.91 & 3.57 & 10.88 & 7.20 \\
 &  & 4 & 2 & 3 & 3 & 16 & 33 & 49 & 16 \\
\cdashline{2-10}
\hline
\multirow{8}{*}{\textbf{JER}} & \multirow{4}{*}{CoNLL-04} & 56.38 & 53.78 & 53.77 & 56.11 & 0.00 & 0.00 & 0.00 & 0.00\\
 &  & 1 & 15 & 2 & 0 & 762 & 632 & 646 & 645\\
 &  & 4.31 & 3.92 & 5.00 & 2.52 & 0.00 & 0.00 & 0.00 & 0.00\\
 &  & 19 & 16 & 19 & 29 & 464 & 611 & 506 & 554\\
\cdashline{2-10}
 & \multirow{4}{*}{NYT} & 79.21 & 81.17 & 80.64 & 80.40 & 77.64 & 82.85 & 82.66 & 82.03\\
 &  & 440 & 224 & 310 & 369 & 1143 & 252 & 363 & 546\\
 &  & 51.94 & 25.46 & 41.13 & 19.76 & 48.18 & 37.27 & 39.00 & 38.39\\  
 &  & 1 & 129 & 17 & 293 & 134 & 188 & 127 & 143\\
\cdashline{2-10}
\hline
\multirow{2}{*}{\textbf{SRL}} & \multirow{2}{*}{CoNLL-12} & 53.68 & 60.35 & 64.48 & 68.34 & 49.40 & 60.39 & 63.50 & 68.33 \\
 &  & 11 & 78 & 3 & 3 & 960 & 1005 & 350 & 159 \\
\cdashline{2-10}
\hline
\multirow{4}{*}{\textbf{ID}} & \multirow{2}{*}{ATIS} & 85.18 & 84.09 & 84.22 & 84.09 & 79.54 & 85.85 & 85.66 & 84.77 \\
 &  & 9 & 12 & 10 & 5 & 121 & 92 & 93 & 94 \\
\cdashline{2-10}
 & \multirow{2}{*}{SNIPS} & 89.02 & 90.44 & 89.87 & 91.16 & 85.37 & 87.80 & 88.33 & 90.17 \\
 &  & 0 & 0 & 0 & 0 & 57 & 8 & 14 & 9 \\
\cdashline{2-10}
\hline
\multicolumn{2}{|c|}{\multirow{2}{*}{\textbf{Average}}} & 46.60 & 46.31 & 47.18 & 46.41 & 38.21 & 41.41 & 42.94 & 43.05 \\
\multicolumn{2}{|c|}{} & 97.50 & 92.25 & 83.42 & 99.83 & 1490.33 & 1783.75 & 1754.83 & 1602.42 \\
\hline
\end{tabular}
\caption{Full result of low resource experiment without structural pre-training (Part 2): SLGM and CE+data. For each dataset, upper row is F1 score, and lower row is format errors.}
\label{tab:low-resource-without-pt-full-part2}
\end{sidewaystable*}

\begin{sidewaystable*}
\centering
\setlength{\tabcolsep}{2pt}
\begin{tabular}{|c|c|c|c|c|c|c|c|c|c|c|c|c|c|}
\hline
\multirow{2}{*}{\textbf{Task}} & \multirow{2}{*}{\textbf{Dataset}} & \multicolumn{4}{c|}{\textbf{CE}} & \multicolumn{4}{c|}{\textbf{CE+FD}} & \multicolumn{4}{c|}{\textbf{CE+FL}} \\ \cline{3-14}
 &  & 0.01 & 0.05 & 0.1 & 0.2 & 0.01 & 0.05 & 0.1 & 0.2 & 0.01 & 0.05 & 0.1 & 0.2 \\ \hline
\multirow{6}{*}{\textbf{NER}} & \multirow{2}{*}{CoNLL-03} & 25.65 & 44.89 & 43.35 & 44.44 & 51.74 & 57.90 & 57.62 & 57.52 & 25.80 & 45.63 & 45.04 & 47.94 \\
 &  & 10480 & 8083 & 7859 & 7781 & 8 & 25 & 21 & 26 & 10184 & 7603 & 7353 & 7420 \\
\cdashline{2-14}
 & \multirow{2}{*}{OntoNotes} & 36.26 & 41.34 & 42.19 & 44.24 & 36.33 & 41.39 & 41.35 & 43.11 & 39.09 & 44.56 & 44.52 & 48.91 \\
 &  & 3108 & 2397 & 2641 & 2741 & 369 & 664 & 452 & 539 & 3884 & 2200 & 2510 & 3069 \\
\cdashline{2-14}
 & \multirow{2}{*}{GENIA} & 45.23 & 51.49 & 53.09 & 48.07 & 41.38 & 45.92 & 47.78 & 45.43 & 47.84 & 53.07 & 55.02 & 54.70 \\
 &  & 528 & 347 & 335 & 300 & 179 & 118 & 88 & 86 & 552 & 372 & 277 & 230 \\
\cdashline{2-14}
\hline
\multirow{4}{*}{\textbf{RE}} & \multirow{2}{*}{TACREV} & 33.68 & 39.19 & 43.60 & 42.50 & 31.84 & 35.56 & 34.53 & 33.29 & 20.42 & 40.41 & 40.98 & 44.76 \\
 &  & 94 & 33 & 91 & 59 & 4 & 3 & 3 & 3 & 118 & 68 & 71 & 100 \\
\cdashline{2-14}
 & \multirow{2}{*}{TACRED} & 38.95 & 44.61 & 45.96 & 44.77 & 38.59 & 44.51 & 46.03 & 44.24 & 39.56 & 45.99 & 46.20 & 45.84 \\
 &  & 597 & 579 & 568 & 588 & 3 & 12 & 12 & 3 & 622 & 73 & 31 & 83 \\
\cdashline{2-14}
\hline
\multirow{8}{*}{\textbf{JER}} & \multirow{4}{*}{CoNLL-04} & 0.00 & 0.00 & 0.00 & 0.00 & 59.66 & 55.38 & 55.57 & 57.02 & 0.00 & 0.00 & 0.00 & 0.00\\
 &  & 758 & 758 & 702 & 756 & 0 & 2 & 1 & 1 & 741 & 710 & 731 & 789\\
 &  & 0.47 & 0.00 & 0.00 & 0.00 & 4.45 & 7.03 & 6.47 & 8.11 & 0.00 & 0.00 & 0.00 & 0.00\\
 &  & 656 & 1037 & 993 & 693 & 208 & 152 & 12 & 12 & 686 & 648 & 593 & 547\\
\cdashline{2-14}
 & \multirow{4}{*}{NYT} & 78.43 & 81.58 & 81.09 & 79.99 & 78.19 & 81.12 & 81.05 & 79.72 & 79.90 & 83.02 & 82.95 & 82.15\\
 &  & 1214 & 417 & 421 & 754 & 744 & 158 & 241 & 442 & 1036 & 326 & 338 & 699\\
 &  & 41.44 & 26.18 & 26.54 & 31.34 & 40.03 & 26.11 & 26.01 & 30.43 & 43.96 & 42.66 & 44.27 & 48.97\\  
 &  & 387 & 347 & 470 & 262 & 61 & 114 & 99 & 60 & 458 & 242 & 216 & 180\\
\cdashline{2-14}
\hline
\multirow{2}{*}{\textbf{SRL}} & \multirow{2}{*}{CoNLL-12} & 52.57 & 61.29 & 64.54 & 68.26 & 51.64 & 61.03 & 64.41 & 67.93 & 55.36 & 63.31 & 66.76 & 70.22 \\
 &  & 1140 & 353 & 230 & 137 & 6 & 5 & 5 & 1 & 1182 & 344 & 229 & 141 \\
\cdashline{2-14}
\hline
\multirow{4}{*}{\textbf{ID}} & \multirow{2}{*}{ATIS} & 82.21 & 84.14 & 83.07 & 83.76 & 83.82 & 83.63 & 82.43 & 82.99 & 83.40 & 84.56 & 84.62 & 85.13 \\
 &  & 124 & 98 & 105 & 95 & 35 & 19 & 20 & 20 & 107 & 85 & 91 & 73 \\
\cdashline{2-14}
 & \multirow{2}{*}{SNIPS} & 85.97 & 91.03 & 90.95 & 91.45 & 86.59 & 90.44 & 90.58 & 91.30 & 88.62 & 92.04 & 92.13 & 91.78 \\
 &  & 50 & 9 & 10 & 11 & 0 & 0 & 0 & 0 & 14 & 7 & 4 & 3 \\
\cdashline{2-14}
\hline
\multicolumn{2}{|c|}{\multirow{2}{*}{\textbf{Average}}} & 43.41 & 47.15 & 47.86 & 48.23 & 50.36 & 52.50 & 52.82 & 53.42 & 43.66 & 49.60 & 50.21 & 51.70 \\
\multicolumn{2}{|c|}{} & 1594.67 & 1204.83 & 1202.08 & 1181.42 & 134.75 & 106.00 & 79.50 & 99.42 & 1632.00 & 1056.50 & 1037.00 & 1111.17 \\
\hline
\end{tabular}
\caption{Full result of low resource experiment with structural pre-training (Part 1): CE, CE+FD, and CE+FL. For each dataset, upper row is F1 score, and lower row is format errors.}
\label{tab:low-resource-with-pt-full-part1}
\end{sidewaystable*}

\begin{sidewaystable*}
\centering
\setlength{\tabcolsep}{2pt}
\begin{tabular}{|c|c|c|c|c|c|c|c|c|c|}
\hline
\multirow{2}{*}{\textbf{Task}} & \multirow{2}{*}{\textbf{Dataset}} & \multicolumn{4}{c|}{\textbf{SLGM}} & \multicolumn{4}{c|}{\textbf{CE+data}} \\ \cline{3-10}
 &  & 0.01 & 0.05 & 0.1 & 0.2 & 0.01 & 0.05 & 0.1 & 0.2 \\ \hline
\multirow{6}{*}{\textbf{NER}} & \multirow{2}{*}{CoNLL-03} & 51.04 & 62.62 & 61.09 & 59.88 & 23.70 & 49.10 & 47.50 & 52.50 \\
 &  & 32 & 28 & 29 & 27 & 10380 & 9046 & 8494 & 7328 \\
\cdashline{2-10}
 & \multirow{2}{*}{OntoNotes} & 38.13 & 45.45 & 44.35 & 46.46 & 35.89 & 45.29 & 44.14 & 46.36 \\
 &  & 594 & 712 & 408 & 733 & 4272 & 8931 & 8713 & 8221 \\
\cdashline{2-10}
 & \multirow{2}{*}{GENIA} & 43.82 & 47.42 & 49.64 & 49.04 & 46.36 & 51.73 & 54.35 & 52.91 \\
 &  & 200 & 122 & 97 & 135 & 594 & 841 & 748 & 476 \\
\cdashline{2-10}
\hline
\multirow{4}{*}{\textbf{RE}} & \multirow{2}{*}{TACREV} & 20.99 & 40.30 & 41.20 & 44.86 & 5.56 & 30.49 & 34.92 & 34.28 \\
 &  & 2 & 2 & 2 & 2 & 37 & 22 & 53 & 35 \\
\cdashline{2-10}
 & \multirow{2}{*}{TACRED} & 38.82 & 45.92 & 46.26 & 46.38 & 28.07 & 30.90 & 25.74 & 21.07 \\
 &  & 5 & 2 & 3 & 3 & 202 & 64 & 36 & 35 \\
\cdashline{2-10}
\hline
\multirow{8}{*}{\textbf{JER}} & \multirow{4}{*}{CoNLL-04} & 61.46 & 57.19 & 58.31 & 58.93 & 0.00 & 0.00 & 0.00 & 0.00\\
 &  & 0 & 0 & 1 & 0 & 749 & 688 & 664 & 654\\
 &  & 3.72 & 5.12 & 6.64 & 7.79 & 0.00 & 0.00 & 0.00 & 0.00\\
 &  & 275 & 54 & 78 & 23 & 463 & 524 & 425 & 511\\
\cdashline{2-10}
 & \multirow{4}{*}{NYT} & 79.53 & 83.09 & 82.86 & 81.93 & 79.35 & 82.34 & 82.33 & 82.07\\
 &  & 461 & 301 & 172 & 470 & 913 & 621 & 565 & 627\\
 &  & 42.64 & 41.54 & 42.81 & 47.67 & 47.58 & 47.69 & 52.20 & 52.71\\  
 &  & 72 & 40 & 25 & 14 & 133 & 185 & 144 & 151\\
\cdashline{2-10}
\hline
\multirow{2}{*}{\textbf{SRL}} & \multirow{2}{*}{CoNLL-12} & 54.29 & 62.98 & 66.61 & 69.95 & 52.49 & 61.58 & 66.04 & 69.87 \\
 &  & 5 & 4 & 2 & 1 & 3008 & 1082 & 467 & 214 \\
\cdashline{2-10}
\hline
\multirow{4}{*}{\textbf{ID}} & \multirow{2}{*}{ATIS} & 83.18 & 83.00 & 83.66 & 83.66 & 80.95 & 86.82 & 84.73 & 87.15 \\
 &  & 24 & 13 & 15 & 15 & 145 & 81 & 79 & 76 \\
\cdashline{2-10}
 & \multirow{2}{*}{SNIPS} & 88.59 & 91.87 & 92.01 & 91.73 & 86.01 & 91.67 & 91.35 & 92.34 \\
 &  & 0 & 0 & 0 & 0 & 32 & 10 & 14 & 5 \\
\cdashline{2-10}
\hline
\multicolumn{2}{|c|}{\multirow{2}{*}{\textbf{Average}}} & 50.52 & 55.54 & 56.29 & 57.36 & 40.50 & 48.13 & 48.61 & 49.27 \\
\multicolumn{2}{|c|}{} & 139.17 & 106.50 & 69.33 & 118.58 & 1744.00 & 1841.25 & 1700.17 & 1527.75 \\
\hline
\end{tabular}
\caption{Full result of low resource experiment with structural pre-training (Part 2): SLGM and CE+data. For each dataset, upper row is F1 score, and lower row is format errors.}
\label{tab:low-resource-with-pt-full-part2}
\end{sidewaystable*}

Tables \ref{tab:low-resource-without-pt-full-part1}, \ref{tab:low-resource-without-pt-full-part2}, \ref{tab:low-resource-with-pt-full-part1}, and \ref{tab:low-resource-with-pt-full-part2} show full result of low resource settings, without and with structural pre-training on single run.
As mentioned earlier, when looking at result of CE+FL and SLGM with NYT dataset in Tables \ref{tab:low-resource-without-pt-full-part1} and \ref{tab:low-resource-without-pt-full-part2}, there were weird explosion of format errors on some dataset fractions, which caused decreased score.
We think this happened because without structural pretraining, model does not have enough structural understanding ability.
With given small examples, the model tried to extract same entities multiple time when sentence goes longer.

\end{document}